\pdfoutput=1

\documentclass[11pt]{article}

\usepackage[final]{acl}

\usepackage{times}
\usepackage{latexsym}
\usepackage{amsmath,amssymb,amsthm}
\usepackage{graphicx}
\usepackage{ulem}
\usepackage{microtype}
\usepackage{hyperref}
\usepackage{url}
\usepackage{booktabs}
\usepackage{colortbl}
\usepackage{array}
\usepackage{lineno}
\usepackage{csquotes}
\usepackage{multirow}
\usepackage{amsfonts}
\usepackage{pifont}
\usepackage{enumitem}
\usepackage{tikz}
\usepackage[skins]{tcolorbox}
\usepackage{wrapfig}
\usepackage{subfigure}
\usepackage{titletoc}

\usepackage[T1]{fontenc}

\usepackage[utf8]{inputenc}

\usepackage{microtype}

\usepackage{inconsolata}
\usepackage{enumitem}
\usepackage{graphicx}
\usepackage{titletoc}
\usepackage{xspace}
\usepackage{booktabs} 
\usepackage{multirow}
\usepackage{amsmath}

\title{Understanding the Thinking Process of Reasoning Models: A Perspective from Schoenfeld's Episode Theory}

\author{%
    \textbf{Ming Li*}, 
    \textbf{Nan Zhang*}, 
    \textbf{Chenrui Fan*}, 
    \textbf{Hong Jiao}\\
    \textbf{Yanbin Fu}, 
    \textbf{Sydney Peters}, 
    \textbf{Qingshu Xu},
    \textbf{Robert Lissitz},
    \textbf{Tianyi Zhou}\\
    {University of Maryland}\\
    \texttt{\{minglii,hjiao\}@umd.edu}, \texttt{tianyidavidzhou@gmail.com} \\
    Project: \url{https://github.com/MingLiiii/Schoenfeld_Reasoning}
}

\begin{document}
\maketitle

\renewcommand{\thefootnote}{}
\footnotetext{*Co-First Authors. }
\renewcommand{\thefootnote}{\arabic{footnote}}

\begin{abstract}	
While Large Reasoning Models (LRMs) generate extensive chain-of-thought reasoning, we lack a principled framework for understanding how these thoughts are structured. In this paper, we introduce a novel approach by applying Schoenfeld's Episode Theory, a classic cognitive framework for human mathematical problem-solving, to analyze the reasoning traces of LRMs. We annotated thousands of sentences and paragraphs from model-generated solutions to math problems using seven cognitive labels (e.g., Plan, Implement, Verify). The result is the first publicly available benchmark for the fine-grained analysis of machine reasoning, including a large annotated corpus and detailed annotation guidebooks. Our preliminary analysis reveals distinct patterns in LRM reasoning, such as the transition dynamics between cognitive states. This framework provides a theoretically grounded methodology for interpreting LRM cognition and enables future work on more controllable and transparent reasoning systems.
\end{abstract}

\section{Introduction}

Large Reasoning Models (LRMs) such as OpenAI GPT-o1 \citep{openai2024o1} and the open-source DeepSeek-R1 \citep{deepseekai2025deepseekr1incentivizingreasoningcapability} exemplify a shift toward producing long, explicit, chain-of-thought (CoT) \cite{wei2023chainofthoughtpromptingelicitsreasoning} that boost performance on demanding tasks \citep{xiong2025selfrewardingcorrectionmathematicalreasoning, xia2025evaluatingmathematicalreasoningaccuracy, liu2024codemindframeworkchallengelarge, wang2023geminireasoningunveilingcommonsense, shao2024deepseekmathpushinglimitsmathematical,xu2024surveyknowledgedistillationlarge}.
These extended and fine-grained reasoning trajectories emerge either from reinforcement-learning optimization \citep{deepseekai2025deepseekr1incentivizingreasoningcapability} or from supervised fine-tuning on expert high-quality responses \citep{cherry,Li2024SuperfilteringWD, li2025instruction, ye2025limoreasoning, muennighoff2025s1simpletesttimescaling}, showcasing increasingly sophisticated thinking patterns.

However, we still lack a principled understanding of how these models organize their problem-solving process. \textit{Are the thinking behaviors of these advanced LRMs similar to Humans? Can we utilize the existing cognitive theories for humans to analyze the LRM reasoning behaviors?} Some researchers notice the extremely human-like thinking token patterns such as ``Hmmm.'', ``Wait.'', ``Let me check.'', ``Am I correct?'' \citep{mitchell2025ai, zhao2025tradeoffs, fan2025missing, li2025makes}, and human-aligned meta-cognitive patterns \citep{steyvers2025metacognition, huang2025beyond}. In a more quantitative way, \citet{shan2025cognitivememorylargelanguage} and \citet{musker2025llmsmodelsanalogicalreasoning} utilized cognitive views on analyzing the thinking process of LRMs.
However, most of them are from observation and summary rather than grounded in solid theories \cite{gandhi2025cognitive, marjanovic2025deepseekr1}.

\begin{table*}[t]
    \centering
    \resizebox{0.95\textwidth}{!}{%
    \begin{tabular}{
        m{0.15\textwidth}
        m{0.85\textwidth}
    }
    \toprule
    \textbf{Category} & \textbf{Reasoning Trace Example (Sentence)}\\
    \midrule
    Read & ``The question asks us to find the value of x in the equation \(2x + 5 = 10\).''\\ 
    \midrule
    Analyze & ``According to the Pythagorean theorem, the square of the hypotenuse is equal to ...'' \\ 
    \midrule
    Plan & ``Next, we will differentiate both sides of the equation with respect to x.''\\ 
    \midrule
    Implement & ``Substituting \(x = 3\) into the equation, we get \(2(3) + 5 = 6 + 5 = 11\).'' \\ 
    \midrule
    Explore & ``Maybe we can try substituting different values for x to see if we can find a pattern.''\\ 
    \midrule
    Verify & ``Let me double-check my calculations: ... which matches the previous result.'' \\ 
    \midrule
    Monitor & ``But wait, hold on.''\\ 
    \bottomrule
    \end{tabular}
    }
    \caption{Representative examples for each episode category in the reasoning traces of LRMs.}
    \label{tab:example}
\end{table*}

Motivated by the advanced theoretical analysis on human behaviors in the area of cognitive science, we propose to utilize the \textit{Schoenfeld's Episode Theory} \citep{schoenfeld2014mathematical} as an analytical framework to understand the thinking process of LRMs. 
This theory was built on hundreds of hours of recorded tape of students tackling non-routine math problems while being asked to think aloud. 
Widely regarded as a gold-standard framework in mathematics-education research, this theory offers a rigorously validated, fine-grained lens for dissecting both expert and novice problem-solving strategies.
After thorough investigation, we find that the thinking process of LRMs can be well-aligned with the episodes in the theory: the $7$ fine-grained episode categories, including \textit{Read}, \textit{Analyze}, \textit{Plan}, \textit{Implement}, \textit{Explore}, \textit{Verify}, and \textit{Monitor}, are also presented in the reasoning traces of LRMs, as they also follow similar problem-solving processes. Examples for each episode are shown in Table \ref{tab:example}.

In this paper, we apply Schoenfeld's Episode Theory to the reasoning traces generated by one of the most representative open-sourced LRMs, DeepSeek-R1 \cite{deepseekai2025deepseekr1incentivizingreasoningcapability}, for both paragraph-level and sentence-level annotation and analysis. 
Specifically, we collect R1 responses on $1,385$ SAT Mathematics items retrieved from the SAT\textregistered  Suite Question Bank\footnote{https://satsuitequestionbank.collegeboard.org/}, as it includes additional meta information, including the difficulty level and the problem domain. 
Then the reasoning responses are annotated and analyzed at both paragraph and sentence levels.

Specifically, in our annotation system, there are $3$ general categories at the paragraph level, including \textit{General}, \textit{Verify}, \textit{Explore}, and $7$ fine-grained episode categories at the sentence level. 
The annotation procedure consists of two parts: 
Firstly, human annotators are trained to manually annotate the responses into proper categories, until the inter-rater reliability reaches a certain level.
Secondly, all the generated reasoning traces are annotated by the trained annotators. 
In total, $38$ math problems, including $915$ paragraphs and $3,087$ sentences, are mannuly annotated into episode categories. 
We also leverage LLM-based and SLM-based methods for labeling and evaluate their efficiency and consistency compared to human labeling, not only formulating this as a well-defined task, but also providing a replacement for the labor-intensive annotation process, making it a more scalable analytical framework for analyzing the reasoning process of LRMs. 
Our contributions are summarized:
\begin{itemize}[leftmargin=0.5em]
    \item We propose the earliest exploration on applying Schoenfeld's Episode Theory to reasoning traces of LRMs, providing a unified view between how humans and LRMs solve math problems. 
    \item We provide a theoretically grounded analytical framework for understanding and analyzing LRM thinking process from a cognitive view.
    \item We release the open annotation protocol and annotated corpus with thousands of annotations for the above-mentioned task, making it a well-defined task for controllable analysis. 
\end{itemize}

\section{Cognitive Foundations}

\subsection{Large Reasoning Models}
LRMs differ from earlier instruction-following LLMs \cite{Scao2022BLOOMA1, mishra2021cross, wei2022finetuned, Chung2022ScalingIL, openai2023gpt4, touvron2023llama} by dedicating both their training and inference budgets to explicit chain-of-thought computation. 
For instance, GPT-o1 \cite{openai_o1_2024} is trained with reinforcement learning that rewards intermediate reasoning traces and is allowed to ``think longer'' at inference, which means that the thinking time and output are both lengthened. This steadily improves accuracy as additional test-time compute is spent. 
Following a similar philosophy, DeepSeek-R1 \citep{deepseekai2025deepseekr1incentivizingreasoningcapability} directly optimizes a reward that incentivizes logically coherent multi-step solutions, yielding strong performance. 
These designs contrast with earlier instruction-following models such as GPT-4o, whose inference paths remain short and lengths relatively fixed. 
Based on the open-sourced thinking traces provided by R1, the communities have noticed the flexible response lengths of R1 when targeting different problems, and the human-like thinking token patterns on the explicit textual thinking traces. 
These human-like behaviors and human-readable thinking traces during problem-solving make it possible to apply behavioral analysis from a cognitive perspective in human research. 

\subsection{Schoenfeld's Episode Theory}

Schoenfeld's Episode Theory \citep{schoenfeld2014mathematical,schoenfeld2016learning} frames problem-solving as a temporally ordered sequence of ``episodes'' that reveal both the solver's evolving goal structure and the meta-cognitive control.
This theory was built on hundreds of hours of recorded tape of students tackling non-routine math problems while being asked to think aloud. 
The original episodes include $6$ categories, i.e., \textit{Read}, \textit{Analyze}, \textit{Plan}, \textit{Implement}, \textit{Explore}, and \textit{Verify}. 
Crucially, the theory disentangles what knowledge a solver possesses from how it is strategically deployed, emphasizing that expert performance hinges less on sheer domain knowledge than on the dynamic orchestration of planning, monitoring, and evaluation processes. 
Episode theory has since become a foundational analytic lens in mathematics-education research and in studies of human reasoning, offering a fine-grained vocabulary for tracing cognitive control and strategy shifts \citep{harskamp2007schoenfeld}.
All of the above characteristics are directly pertinent when we scrutinize the reasoning traces generated by contemporary large language models.
See Appendix \ref{appendix:theories} for a more detailed discussion and comparison on different cognitive theories used in human research.

\subsection{Why Episode Theory Fits LRM Traces}

Schoenfeld's theory represents a natural framework for understanding problem-solving processes, as it was originally developed to capture how humans naturally approach and resolve mathematical challenges. 
The theory encapsulates the organic flow of cognitive processes that occur when individuals encounter and work through math problems.
After through investigation on LRM generated reasoning traces, we actually find their reasoning structures are well-aligned with the theory:
Typically, when facing a given problem, the models will \textit{Read} and restate the problem in a form that they can understand better, then \textit{Analyze} the potential strategies to solve the problem. After that, they will \textit{Plan} for the future steps right before they \textit{Implement} the calculation. Sometimes, when the problem cannot be easily solved, LRMs will \textit{Explore} other potential methods and \textit{Verify} their generated traces. 
Among these categories, \textit{Explore} and \textit{Verify} further represent the current traned on test-time scaling and the occurrence of the ``aha moment'' \citep{deepseekai2025deepseekr1incentivizingreasoningcapability} of LRMs. 

Moreover, since LRMs externalize these transitions between episodes in natural-language form, which shows a better form for further analysis than humans. 
LRM's long reasoning traces can be segmented clearly through paragraph-level and sentence-level boundaries, and then annotated at the same fine-grained episode categories used to analyze human problem-solving sessions. 
This process uncovers the system's dynamic metacognitive regulation rather than leaving it a black box. 
Consequently, Episode Theory offers a uniquely well-suited lens for analyzing the LRM reasoning behavior.
Representative examples for each episode category are shown in Table \ref{tab:example}, and an fully annotated reasoning trace is shown in Appendix \ref{appendix:example}.



\section{Dataset Construction}

\subsection{Data Source}
Our experiments are conducted on $1,385$ SAT Mathematics items retrieved from the SAT\textregistered  Suite Question Bank, which contains $19$ fine-grained skill categories. 
SAT serves as a nationwide college admission test in the U.S., which is usually taken by junior and senior high school students. 
There are two types of questions in SAT Math: multiple-choice (MC) items and student-produced response (SPR) items. 
Each item consists of the following parts of information: question text, skill, difficulty level, and a step-by-step human solution (rationale). Depending on the question type, choices, figures, or tables could exist.
For LRM-generated responses, DeepSeek-R1 is selected since its thinking trajectories can be obtained during the inference. 
Specifically, $38$ responses, $2$ for each fine-grained skill category, including $915$ paragraphs, and $3,087$ sentences, are annotated.

\subsection{Data Annotation}

\paragraph{Annotation Strategy.}
We leverage Schoenfeld's Episode Theory to systematically annotate the reasoning processes of LRMs. Compared to the traditional ``think aloud'' protocols used for human problem-solving, LRM-generated responses often exhibit much finer granularity, with explicit steps and monitoring indicators articulated in the output. This increased granularity can sometimes make it challenging to assign clear episode labels to every sentence. For example, in some cases, all several paragraphs in the response may be related to a certain behavior like \textit{Verify}. However, during the whole verify process, the model might still conduct more finegrained behaviors, such as \textit{Plan}, (plan on how to verify), etc. Thus, a better annotation strategy is required to tackle these issues like ``Plan during Verify''. 

To address this, we adopt a hierarchical annotation strategy that operates at both the paragraph and sentence levels. At the paragraph level, we capture broader, long-term behaviors, such as the initial attempt to solve a problem or the process of verifying an existing answer. At the sentence level, we annotate more fine-grained strategies, for example, the specific planning, implementation, or verification steps that occur within a broader verification episode. By leveraging contextual information and this hierarchical approach, we are able to define and label sentences more precisely, even in cases where boundaries between categories are subtle. 

For paragraph-level annotation, each response is segmented into paragraphs following the original formatting. Each paragraph is then assigned one of three labels, \textit{General} if the paragraph is directly solving the math problem, \textit{Explore} if the paragraph diverges from the main solution to investigate possibilities or gather insights, or \textit{Verify} if the paragraph is checking a solution, according to the dominant episode it represented. 
For sentence-level annotation, we further split each response to isolate individual sentences. Afterward, each sentence is labeled into one of the seven episodes: \textit{Read} if the sentence is repeating the question; \textit{Analyze} if the sentence is recalling relevant theories, deducing relationships, or introducing symbols; \textit{Plan} if the sentence is announcing the next step; \textit{Implement} if the sentence is executing a planned strategy; \textit{Explore} if the sentence is generating potential ideas or making guesses; \textit{Verify} if the sentence is judging the correctness. An extra category \textit{Monitor} was added to serve as transitions between episodes, such as self-monitoring or hesitation. Examples for each episode are shown in Table \ref{tab:example}.
Since paragraphs and sentences form an interwoven hierarchy, annotators work in sequence for each item: they first label all paragraphs, ensuring coherent episode-level categorization, and then label each extracted sentence, capturing finer moves within those episodes.

\paragraph{Annotation Process.}
We first develop a preliminary annotation guidebook grounded in Schoenfeld's Episode Theory. Next, two of the authors each annotate $5$ R1 responses using the initial guide. This pilot annotation familiarizes the annotators with the structure of the responses and highlights areas of ambiguity or inconsistency in the guide itself. Based on their annotation experience, the annotation definitions and examples are further refined. After the guidebook is determined, three annotators are trained and complete all the reamining responses. See Appendix \ref{appendix:codebook_para} and Appendix \ref{appendix:codebook_sent} for the full guidebooks. 
An annotated example is shown in Appendix \ref{appendix:example}.

\begin{table*}[t]
  \centering
  \begin{tabular}{l|*{8}{c}}
      \toprule
      \multirow{2}{*}{\textbf{Model}} &
      \multicolumn{2}{c}{\textbf{Base}} &
      \multicolumn{2}{c}{\textbf{Example}} &
      \multicolumn{2}{c}{\textbf{Guidebook}} &
      \multicolumn{2}{c}{\textbf{Ex+Guide}} \\
      \cmidrule(lr){2-3}\cmidrule(lr){4-5}\cmidrule(lr){6-7}\cmidrule(lr){8-9}
      & \textbf{Para} & \textbf{Sent} & \textbf{Para} & \textbf{Sent} &
        \textbf{Para} & \textbf{Sent} & \textbf{Para} & \textbf{Sent} \\
      \midrule
      \textbf{GPT-4.1}    & 0.444 & \textbf{0.595} & 0.559 & 0.604 & \textbf{0.740} & \textbf{0.676} & \textbf{0.757} & \textbf{0.681} \\
      \textbf{Gemini-2.0} & \textbf{0.460} & 0.590 & \textbf{0.647} & \textbf{0.628} & 0.723 & 0.655 & 0.697 & 0.626 \\
      \textbf{GPT-4o}     & 0.388 & 0.475 & 0.537 & 0.504 & 0.656 & 0.577 & 0.714 & 0.609 \\
      \bottomrule
  \end{tabular}
  \caption{Comparison of paragraph-level accuracy (Para) and sentence-level accuracy (Sent) across models with different prompting techniques. \textit{Base} uses zero-shot prompting only; \textit{Example} provides ground-truth in-context examples; \textit{Guidebook} adds our detailed guidebook; \textit{Ex+Guide} combines both methods. The performance shows the extraordinary performance improvement of using our detailed guidebook for automatic annotation.}
  \label{tab:llm_acc}
\end{table*}

\begin{table}[t]
    \centering
    \resizebox{0.48\textwidth}{!}{%
    \begin{tabular}{c}
        \begin{tabular}[t]{lccccccc}
            \toprule
            \multicolumn{8}{c}{\textbf{GPT-4.1}}\\
            \cmidrule(lr){2-8}
            & \textbf{Read} & \textbf{Analyze} & \textbf{Plan} & \textbf{Impl.} & \textbf{Expl.} & \textbf{Verif.} & \textbf{Monit.}\\
            \midrule
            \textbf{Read} & 11.3 & 0.1 & 0.0 & 0.0 & 0.1 & 0.5 & 0.2\\
            \textbf{Analyze} & 0.8 & 16.6 & 0.3 & 0.2 & 2.3 & 4.8 & 0.6\\
            \textbf{Plan} & 0.1 & 0.0 & 5.7 & 0.1 & 0.2 & 0.5 & 0.6\\
            \textbf{Implement} & 0.0 & 0.5 & 0.5 & 17.1 & 0.4 & 3.2 & 0.1\\
            \textbf{Explore} & 0.0 & 0.0 & 0.1 & 0.0 & 6.1 & 0.4 & 0.1\\
            \textbf{Verify} & 0.0 & 0.4 & 0.1 & 0.3 & 0.3 & 18.5 & 0.7\\
            \textbf{Monitor} & 0.0 & 0.0 & 0.1 & 0.0 & 0.2 & 0.5 & 5.2\\
            \bottomrule
        \end{tabular}\\[0.8em]
        \begin{tabular}[t]{lccccccc}
            \toprule
            \multicolumn{8}{c}{\textbf{BERT}}\\
            \cmidrule(lr){2-8}
            & \textbf{Read} & \textbf{Analyze} & \textbf{Plan} & \textbf{Impl.} & \textbf{Expl.} & \textbf{Verif.} & \textbf{Monit.}\\
            \midrule
            \textbf{Read} & 9.0 & 2.6 & 0.2 & 0.2 & 0.2 & 0.1 & 0.0\\
            \textbf{Analyze} & 1.2 & 19.8 & 1.0 & 0.8 & 1.1 & 2.5 & 0.1\\
            \textbf{Plan} & 0.9 & 0.8 & 4.3 & 0.9 & 0.1 & 0.3 & 0.2\\
            \textbf{Implement} & 0.2 & 3.5 & 0.5 & 16.5 & 0.1 & 1.1 & 0.5\\
            \textbf{Explore} & 0.2 & 0.6 & 0.2 & 0.2 & 4.9 & 0.8 & 0.2\\
            \textbf{Verify} & 0.1 & 2.5 & 0.5 & 0.9 & 0.5 & 13.6 & 0.3\\
            \textbf{Monitor} & 0.0 & 0.2 & 0.3 & 0.0 & 0.1 & 1.0 & 4.2\\
            \bottomrule
        \end{tabular}
    \end{tabular}%
    }
    \caption{Sentence-level confusion matrices (percentage) for GPT-4.1 (top) and BERT (bottom). Rows correspond to true labels and columns to predicted labels.}
    \label{tab:confusion}
\end{table}

\begin{table}[h]
    \centering
    \begin{tabular}{lcc}
        \toprule
        \textbf{Model} & \textbf{Accuracy} & \textbf{Cohen’s $\kappa$} \\
        \midrule
        \textbf{GPT-4.1 (Ex+Guide)} & \textbf{0.805} & \textbf{0.764} \\
        \midrule
        \textbf{BERT}                    & 0.732 & 0.671 \\
        \textbf{RoBERTa}                 & 0.730 & 0.670 \\
        \textbf{SVM‐Gemini}        & 0.704 & 0.632 \\
        \textbf{MLP‐Gemini}              & 0.684 & 0.613 \\
        \textbf{KNN‐Gemini}            & 0.587 & 0.490 \\
        \bottomrule
    \end{tabular}
    \caption{Overall sentence-level accuracy and Cohen’s $\kappa$ for each model on the $30\%$ test subset. GPT-4.1 obtains the best results, while for training-based methods, the BERT model obtains the best performance. }
    \label{tab:slm_acc}
\end{table}

\section{Experiments and Analysis}

In this section, we explore the potential automated annotation methods for our task.

\begin{figure}[t]
    \centering
    \includegraphics[width=1.0\columnwidth]{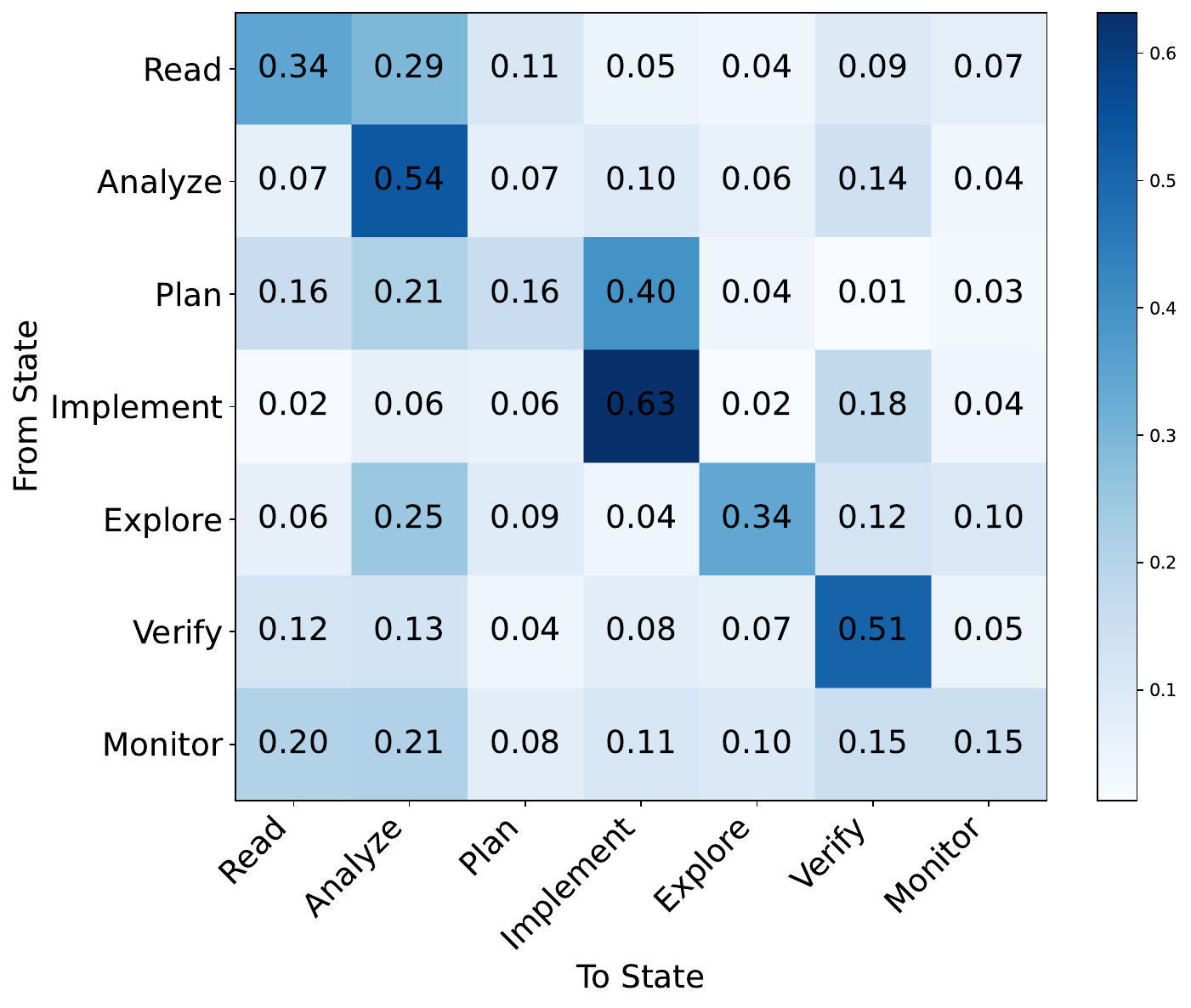}
    \caption{The sentence-level state transition matrix for our ground-truth annotation. A darker color represents a higher probability of transfer from one state to another.   }
    \label{fig:sent_trans}
\end{figure}

\subsection{Transition Matrix}


Figure \ref{fig:sent_trans} presents the state transition matrix for our ground-truth annotation. A darker color represents a higher probability of transfer from one state to another. The darkest values aside from the diagonal are \textit{Read-Analyze}, \textit{Plan-Implement}, and \textit{Explore-Analyze}, representing that these categories have a higher probability of appearing continuously. For example, \textit{Plan-Implement} represents that \textit{Implement} has more chance to follow the \textit{Plan}. This matrix directly shows the relation between different categories, which further shows a strong alignment with human behaviors. 

\subsection{LLM-based Zero-shot Methods}

In this section, we investigate how advanced LLMs perform on Schoenfeld’s Episode Theory annotations as shown in Table \ref{tab:llm_acc}. In the table, we present the results of using \textit{GPT-4.1}, \textit{GPT-4o}, and \textit{Gemini-2.0-flash} on different prompting techniques. \textit{Base} represents using LLM zero-shot prompting only. \textit{Guidebook} represents using our detailed guidebook as the guidance. \textit{Example} represents providing some ground-truth examples for in-context learning. \textit{Ex+Guide} represents using both of the mentioned methods. These experiments are conducted on all of our annotated data. 
The performance shows the extraordinary performance improvement of using our detailed guidebook for automatic annotation. Comparing the performances of different LLMs, we notice that GPT-4.1 can obtain the best performance most of the time. Gemini-2.0 can outperform others in the in-context learning scenarios.

\subsection{Training-based methods}

For the training-based method, we split all our annotated data into training ($70\%$) and testing ($30\%$) subsets, and finetune BERT-level models, including BERT \cite{devlin-etal-2019-bert} and RoBERTa \cite{liu2019robertarobustlyoptimizedbert}. 
In addition, to investigate how powerful embeddings help the task, we try stronger embedding models, Gemini, and utilize its embeddings to train SVM and MLP for the classification. Based on Gemini embedding, we also try KNN method to see how it performs. 
As shown in Table \ref{tab:slm_acc}, GPT-4.1 obtains the best results, with the highest accuracy ($0.805$) and the highest $\kappa$ ($0.764$). The detailed confusion matrices on sentence-level accuracy on both GPT-4.1 and BERT are shown in Table \ref{tab:confusion}.
According to the confusion matrices, we can find the highest values aside from the diagonal are \textit{Analyze-Verify}, \textit{Implement-Verify}, and \textit{Verify-Implement}, which represent scenarios where even the best model can not perform well.

\section{Conclusion}

In this work, we bridged a gap between cognitive science and artificial intelligence by using Schoenfeld's Episode Theory to analyze the reasoning of Large Reasoning Models. We demonstrated that a framework designed for human problem-solving can effectively decode machine-generated thought processes, revealing a structured, episodic nature in how LRMs tackle mathematical challenges. The core of our contribution is a novel, large-scale annotated corpus and a reusable analytical protocol, which we have made publicly available to foster further research. This research not only offers initial insights into the thinking patterns of current models but also establishes a foundational methodology for future investigations. 

\clearpage
\section*{Limitations}

The main limitation of this version of the paper is the scope. Currently, only SAT math data is taken into account and labeled. To enhance the dataset's size and complexity, future iterations should incorporate data from additional sources. For example, since the SAT is designed as a college admission test in the U.S., the overall difficulty level is relatively moderate. As a next step, we plan to include other datasets, which are derived from mathematical Olympiad competitions and features more challenging items.
Besides, the accuracy for automatic annotation is not very high, further efforts are needed to improve the accuracy.

\bibliography{custom}

\clearpage

\appendix
\onecolumn
\startcontents[appendix]
\printcontents[appendix]{ }{0}{\section*{Table of Contents for Appendix}}

\clearpage

\section{Theories of Mathematical Problem Solving}
\label{appendix:theories}
Analyzing human problem-solving behaviors has long been central to cognitive psychology and mathematics education, informing theoretical models and annotation practices used extensively in research. Early studies frequently employed broad cognitive taxonomies such as the revised Bloom's Taxonomy \citep{krathwohl2002revision}, which categorizes cognitive processes hierarchically into \textit{Remember}, \textit{Understand}, \textit{Apply}, \textit{Analyze}, \textit{Evaluate}, and \textit{Create}. Although influential for setting educational objectives and evaluating task complexity, Bloom's framework was originally a classification system set for educational learning objectives in the cognitive domain. Therefore, it failed to capture the nuanced cognitive strategies and iterative processes inherent in mathematical problem-solving \citep{darlington2013use, radmehr2018revised} and was inconsistent with our analysis purpose. 

To address these limitations, researchers developed domain-specific frameworks. One seminal model is \citet{polya1945how} four-phase model, consisting of \textit{Understanding the problem}, \textit{Devising a plan}, \textit{Carrying out the plan}, and \textit{Reflecting} (looking back). Pólya's framework influenced both instructional practices and subsequent theoretical developments. Teachers often use this model as an explicit guide to help students structure their problem-solving approach in mathematics classrooms. In terms of coding and annotation of behavior, Pólya's phases provide a general partitioning of the problem-solving timeline. For instance, early studies analyzed students' think-aloud protocols or written solutions by tagging broad segments as “understanding” versus “planning.” However, Pólya's framework was intended as a prescriptive guide and reflective tool, not a fine-grained analytical scheme. It captures the high-level structure of an ideal solving process, but within each phase there can be a lot of cognitive action that Pólya's labels don't distinguish.

Similarly, \citet{mason2010thinking} introduced a three-phase model (i.e., \textit{Entry}, \textit{Attack}, and \textit{Review}) that sought to highlight both cognitive and emotional states during problem-solving. Mason's framework aimed primarily at helping students develop reflective habits in managing their thought processes. Later, this model was refined by \citet{yeo2010characterising}, who provided additional subtasks (such as \textit{Specializing}, \textit{Conjecturing}, and \textit{Generalizing}) within these broader phases and added additional phases \textit{Extension} after the \textit{Review} phase. Despite this enhancement, these models still function mainly as instructional scaffolds, explicitly intended to guide students through the cognitive and metacognitive aspects of problem-solving rather than serving as rigorous empirical coding tools. In other words, while useful for teaching students how to engage effectively with mathematical tasks, their application in systematic behavior annotation and empirical cognitive analysis has been limited, primarily because their detailed subtasks were not fully operationalized or validated for research annotation.

A more refined model explicitly emphasizing cognitive and metacognitive dimensions is \citet{greenes1995mathematics} five-stage framework. Greene proposed a sequential but detailed cognitive approach to problem-solving, consisting of: (1) \textit{Problem representation}, clearly interpreting and formulating the problem's meaning; (2) \textit{Strategy design}, identifying potential solution approaches; (3) \textit{Implementation}, executing the selected strategy or calculation steps; (4) \textit{Monitoring and evaluation}, continuously checking progress and strategy effectiveness; and (5) \textit{Reflection and consolidation}, critically reviewing the final solution and reflecting on problem-solving strategies and outcomes for future problem-solving situations. Greene's model particularly highlighted the critical role of metacognitive monitoring and evaluation, emphasizing that successful problem-solving involves active self-assessment and the willingness to adapt strategies dynamically. However, despite its strength in explicitly addressing metacognitive behaviors, Greene's sequential stage structure can be limiting when coding nonlinear, iterative problem-solving processes typically observed in real-world mathematics tasks.

The most comprehensive and empirically validated framework available is Schoenfeld's episode theory \citep{schoenfeld2014mathematical}. This model codes student behaviors into explicit problem-solving episodes such as \textit{Reading}, \textit{Analysis}, \textit{Exploration}, \textit{Planning}, \textit{Implementation}, and \textit{Verification}. Schoenfeld's episode theory is empirically supported by using detailed protocol analyses, explicitly identifying strategic decisions and critical moments of cognitive transition or failure, and distinguishing successful from unsuccessful problem-solving attempts. By examining transitions between these episodes, Schoenfeld observed that students often failed to solve problems not due to a lack of mathematical knowledge, but because of ineffective metacognitive control. For instance, many students would read the problem and immediately begin exploring solutions without adequate analysis or planning. This premature transition often led them down unproductive paths, highlighting the importance of strategic decision-making at episode boundaries. Therefore, successful problem-solving is heavily influenced by the solver's ability to monitor and regulate their cognitive processes. Further empirical support comes from a study by \citet{kuzle2013patterns}, who investigated the problem-solving behaviors of preservice teachers using dynamic geometry software. Applying Schoenfeld's framework, Kuzle identified that participants who engaged in thorough analysis and planning before implementation were more successful in solving problems. Conversely, those who skipped these critical phases often encountered difficulties, underscoring the role of strategic transitions between episodes in effective problem-solving. Because of its rich granularity and robust empirical grounding, Schoenfeld's model facilitates detailed insights into both cognitive and metacognitive dimensions of mathematical reasoning. 

Recently, as artificial intelligence (AI) systems increasingly produce mathematical explanations, scholars have begun applying human-based frameworks to AI-generated reasoning. While broad frameworks (Bloom's, Pólya's) are limited by their lack of specificity in coding detailed cognitive actions, Schoenfeld's detailed cognitive-metacognitive structure is especially appropriate. His explicit focus on strategic control and self-monitoring aligns closely with known challenges in AI reasoning, such as inadequate self-regulation and verification steps, making his framework uniquely suitable for annotating and analyzing AI-generated responses systematically.

In summary, although broad cognitive frameworks (i.e., Bloom's Taxonomy) and general phase-based models \citep{polya1945how, mason2010thinking, greenes1995mathematics} historically provided foundational instructional value, their limitations become evident when applied to detailed empirical research annotations. Schoenfeld's theory empirically robust, fine-grained cognitive-metacognitive model addresses these limitations, effectively supporting detailed analysis and annotation of both human and AI-generated mathematical reasoning, justifying its selection for the present study.

\clearpage

\section{SAT Data Source}
\label{appendix:DataSource}
The SAT\textregistered Suite Question Bank (https://satsuitequestionbank.collegeboard.org/) math section assesses students' proficiency across four domains, comprising a total of 19 distinct skills. In the Algebra domain, the skills include "Linear equations in one variable," "Linear equations in two variables," "Linear functions," and "Systems of two linear equations." The Advanced Math domain includes "Equivalent expressions," "Nonlinear equations in one variable," "Systems of equations in two variables," "Nonlinear relationships," and "Functions." Within the Problem Solving and Data Analysis domain, the assessed skills are "Ratios, rates, and proportions," "Percents," "Units," "Tables and data inferences," "Data collection and conclusions," and "Probability and conditional probability." Lastly, the Geometry and Trigonometry domain includes "Area and volume," "Lines, angles, and triangles," "Right triangles and trigonometry," and "Circles." These skills collectively reflect the mathematical knowledge and reasoning required for college readiness.

Two types of items are available in SAT math. Multiple-choice (MC) items have four options, with each item having one and only one correct or best answer. Some items on the Math Tests are student-produced response (SPR) items, which require the student to solve a problem and then grid their response on the answer sheet. Each item consists of the following parts of information: Question text, domain, skill, difficulty level, and a step-by-step human solution (rationale). Depending on the question type, choices, figures, or tables could exist. All of the information was collected from the SAT Suite Question Bank and put into a JSON file. The JSON file has the following fields: Question ID, Assessment, Test, Domain, Skill, Question Difficulty, Item Stem, Question, Choice A, Choice B, Choice C, Choice D, Correct Answer Rationale, Table, Figure. Assessment means the specific assessment from the SAT Suite; test specifies the subject of the assessment; domain and skill refer to the aspect of math knowledge tested by each item, which was specified in the test specification of SAT; question difficulty is a three-level categorical variable (i.e., easy, medium, hard), of which the calculation was provided in Appendix; Item Stem is the content of the math problem without the question at the end, which was put in the Question field; all other json fields represent what their names mean. Note that we converted the math language into LaTeX format for the machine to understand.


\clearpage
\section{Example}
\label{appendix:example}

\begin{figure}[h]
    \centering
    \includegraphics[width=0.84\linewidth]{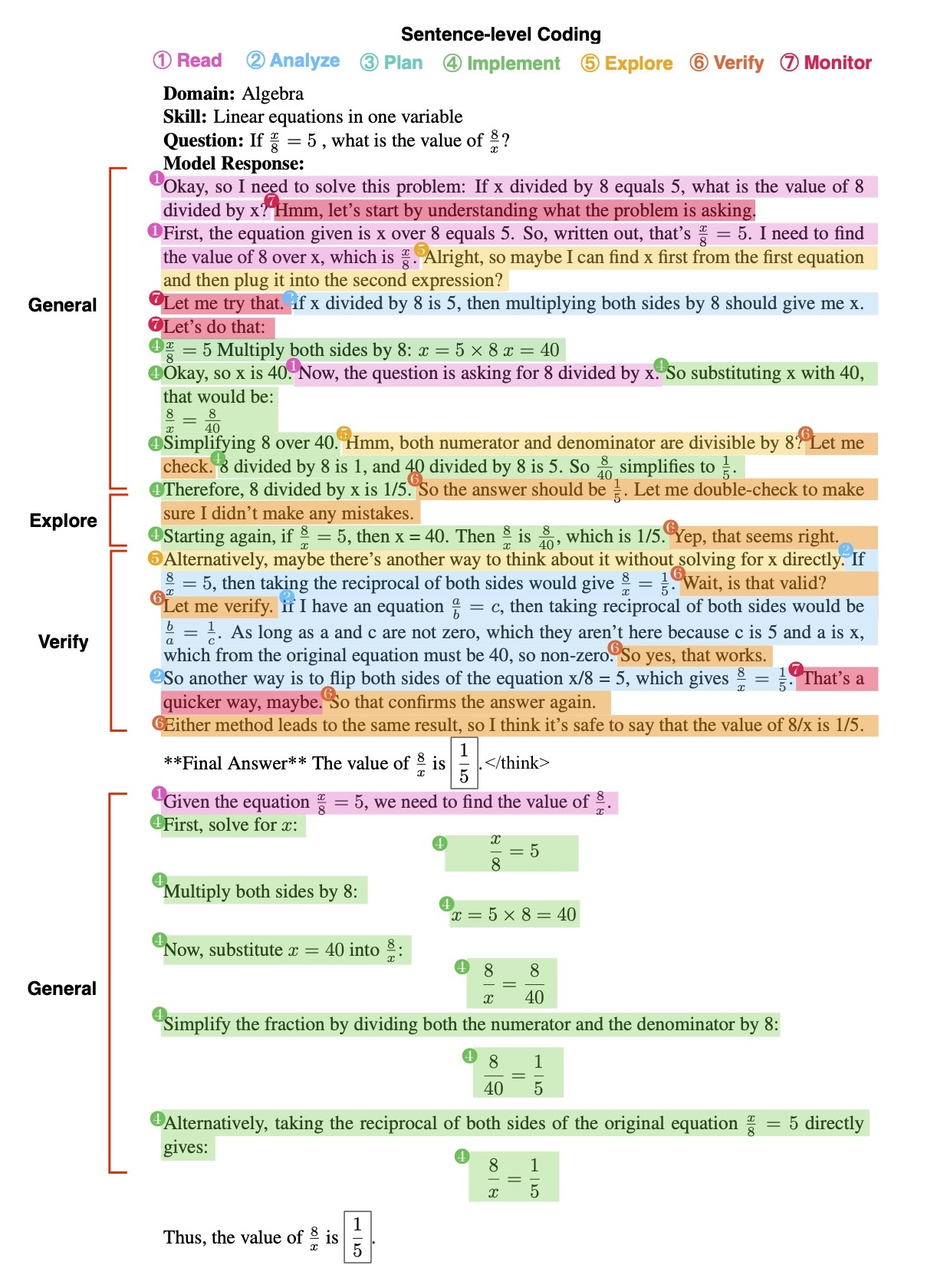}
    \caption{This figure presents an example of an annotated reasoning process under the adapted Schoenfeld's Episode Theory. Paragraph-level annotations are indicated by square brackets on the left, while sentence-level annotations are color-coded by cognitive process categories.}
    \label{fig:example}
\end{figure}

In this section, we present an example of an annotated reasoning process under the adapted Schoenfeld's Episode Theory as shown in Figure \ref{fig:example}. Paragraph-level annotations are indicated by square brackets on the left, while sentence-level annotations are color-coded by cognitive process categories.

\clearpage
\section{\textbf{Paragraph-level Annotation Guidebook}}
\label{appendix:codebook_para}

In this project, we aim to analyze the reasoning process of current large language models (LLMs) with advanced reasoning capabilities, i.e., Large Reasoning Models, LRMs, based on a modified version of Alan Schoenfeld's (1985) "Episode-Timeline" framework for problem-solving. The original Schoenfeld's theory was built on hundreds of hours of recorded tapes of students tackling non-routine math problems while being asked to think aloud. Widely regarded as a gold-standard framework in mathematics-education research, this theory offers a rigorously validated, fine-grained lens for dissecting both expert and novice problem-solving strategies. After thorough investigation, we find that the thinking process of LRMs can be well-aligned with the episodes in the theory, as they also follow similar problem-solving processes. Thus, in this project, we aim to annotate the model (solver) responses with these episode categories. However, to better apply the theory to the analysis of model responses, we apply a hierarchical annotation structure, including paragraph-level annotation and sentence-level annotation. Paragraph-level annotation is used to capture the overall flow of the model response, including \textbf{three} categories: General, Explore, and Verify. Sentence-level annotation is used to capture the fine-grained behavior of each sentence, including \textbf{seven} categories: Read, Analyze, Plan, Implement, Explore, Verify, and Monitor.

This guidebook provides guidelines for \textbf{paragraph-level annotation}. For the given model response paragraphs, each of them will be annotated as one of three broad episode types: General, Explore, or Verify. Please remember, the paragraph-level annotation is used to capture the overall flow of the model response, thus we should take the context into consideration to decide the correct category.

\subsection*{\textbf{Label Definitions and Guidelines}}

\subsubsection*{\textbf{1. General}}

\begin{itemize}[leftmargin=*]
    \item \textbf{Definition:} A General episode is a response paragraph where the solver carries out the initial attempt and main line of problem-solving. This category encompasses all general problem-solving actions that are NOT explicitly exploratory trials (Explore) or verification checks (Verify). It typically includes understanding the problem, devising and following a plan, executing solution steps, and making transitions between ideas in a coherent way. In essence, whenever the response is proceeding methodically toward solving the problem (stating facts, applying formulas, deriving results) without digressing into off-plan explorations or checking of results, it falls under a General episode.
    \item \textbf{Guidelines:}
    \begin{itemize}[leftmargin=*]
        \item Assign the General label to any paragraph that focuses on \textit{direct progress} toward the solution using established methods or known information. And that progress should not be a part of the verification or exploration process.
        \item This can range from restating the problem in the solver's own words, setting up equations, systematically carrying out computations, to explaining a step-by-step plan. Such paragraphs often appear at the beginning of a solution (interpreting or summarizing the problem), throughout the middle as the solver works through solution steps, or even as concluding statements that present the final answer without explicitly verifying it.
        \item Typically, for the solver response, only the paragraphs for the initial attempt belong to the General category. After the initial attempt, the following paragraphs should be annotated as Explore or Verify.
        \item When differentiating General from other categories, focus on the paragraph's purpose. Given that one paragraph contains several sentences, the General episode might contain sentences that occur from time to time which are considered exploration or verification at the sentence level. However, that exploration or verification is still considered embedded in the General chunk, therefore should be General.
    \end{itemize}
\end{itemize}

\subsubsection*{\textbf{2. Explore}}

\begin{itemize}[leftmargin=*]
    \item \textbf{Definition:} An Explore episode is a response paragraph where the solver diverges from the main solution path to investigate possibilities, gather insight, or attempt an approach that isn't guaranteed to work. In Schoenfeld's framework, exploration is a less structured, trial-and-error phase: the solver searches for relevant information or patterns and considers various options without a firm plan. Within a solver response, an Explore paragraph typically reflects the solver's uncertainty or creativity—trying a conjecture, exploring a sub-problem, or experimenting with different strategies. This episode serves to advance understanding when the direct path is unclear, and it often precedes or informs a return to structured problem-solving.
    \item \textbf{Guidelines:}
    \begin{itemize}[leftmargin=*]
        \item Use the Explore label for paragraphs where the reasoning is characterized by searching or experimentation rather than executing a known strategy. If the solver poses hypotheses, tests special cases, or temporarily pursues a tangent, that paragraph should be marked Explore.
        \item Common scenarios include: trying a simpler instance of the problem to see how it behaves, guessing a pattern or formula and seeing if it holds, discussing alternative approaches, or brainstorming next steps when stuck. Watch for a \textit{change in tone}—if the paragraph shifts from solving to musing or investigating, it likely represents an exploratory episode.
        \item Since this annotation is at the paragraph level, please only mark the large chunk of an exploration episode toward the solution of the whole question as Explore. If the exploration is only for one step of the solution, it is Explore at the sentence level, therefore not Explore at the paragraph level. For example, in the middle of solving the problem, there is a step for solving 3x=6, which is only a small part of solving the whole problem. In the solution, people explored two different ways of solving the equation, then these are Explore at the sentence level rather than at the paragraph level. Therefore, the tag should be decided carefully depending on whether these steps are a procedure belonging to the initial solution (General), exploration of a new method (Explore), or verification of the answer (Verify).
        \item To tell an Explore episode apart from General and Verify, focus on purpose and tone. An Explore paragraph is driven by \textit{uncertainty}: it does not guarantee the solution, but is proactive in providing a new approach to think about or to solve the question. Unlike a General episode, which follows a clear logical step in solving, an Explore episode might temporarily set the main goal aside to answer a sub-question or try a different angle. If the paragraph suggests that the solver is unsure how to proceed and is examining options, that's a strong sign of exploration rather than general solving. Also, Explore is distinct from Verify because it's not primarily about checking a result's correctness. Verify episodes confirm what's believed to be a solution, whereas Explore episodes occur before a solution is confirmed (often before one is even reached). In sum, if the paragraph's content is exploratory (testing, conjecturing, or considering alternatives) and doesn't simply carry out a decided plan or finalize an answer, it should be labeled Explore.
    \end{itemize}
\end{itemize}

\subsubsection*{\textbf{3. Verify}}

\begin{itemize}[leftmargin=*]
    \item \textbf{Definition:} A Verify episode is a response paragraph dedicated to confirming the correctness or validity of a solution or a result. In Schoenfeld's terms, a verification episode occurs when the solver "reviews and tests" a solution to ensure it meets the problem requirements. In the context of AI-generated responses, this usually means the paragraph is focused on checking an answer (or an intermediate result) and making sure it is consistent with the conditions of the problem. This episode type often comes at the end of a solution as a final confirmation, but it can also appear mid-solution if the solver pauses to validate a step. The hallmark of a Verify episode is that its primary goal is evaluation: double-checking calculations, substituting the found solution back into the original problem, or logically arguing that the result must be correct.
    \item \textbf{Guidelines:}
    \begin{itemize}[leftmargin=*]
        \item Label a paragraph as Verify when the content is centered on checking, confirming, or proving that a result is correct. If the solver explicitly performs a check (for example, plugging the solution back into the original equation to see if it holds, or verifying that all conditions are satisfied), that paragraph should be tagged as Verify.
        \item Do not label a paragraph as Verify if it is merely stating the final answer without checking it; in that case use General for a straightforward conclusion. Only use Verify when actual validation or reasoning about correctness is present.
        \item When a solver launches a \textit{verification routine} (e.g., "To verify…"), every paragraph that belongs to that routine inherits the Verify label, even if later lines contain no overt "checking" words. Mechanical algebra steps, intermediate equations, and the final boxed answer are all part of the same confirmation process, so they are not General.
        \item The key feature that distinguishes a Verify episode is its retrospective nature: it looks back at a result and asks "is this right?" rather than pushing forward to solve something new. This sets it apart from a General episode, which would simply present the result or next step without questioning it. If a paragraph is primarily re-examining what's been done (whether the final answer or a prior step), it is likely to be verification. Compared to an Explore episode, which is uncertain and looking for a way forward, a Verify episode assumes a candidate solution or fact and checks its validity—it is usually certain about what needs to be checked, just not about whether it passes the check. Another way to distinguish Verify is by position and content: a paragraph that comes after an answer has been derived and discusses that answer in light of the problem's conditions is almost always a Verify episode. Be careful not to confuse a final declarative answer (General) with a verification; if there's no evidence of a \textit{checking process} or confirmation language, then it's not truly a Verify episode. The presence of phrases explicitly referencing correctness or fulfilling conditions is the clearest differentiator for Verify.
    \end{itemize}
\end{itemize}

\clearpage
\section{\textbf{Sentence-level Annotation Guidebook}}
\label{appendix:codebook_sent}

In this project, we aim to analyze the reasoning process of current large language models (LLMs) with advanced reasoning capabilities, i.e., Large Reasoning Models, LRMs, based on a modified version of Alan Schoenfeld's (1985) "Episode-Timeline" framework for problem-solving. The original Schoenfeld's theory was built on hundreds of hours of recorded tapes of students tackling non-routine math problems while being asked to think aloud. Widely regarded as a gold-standard framework in mathematics-education research, this theory offers a rigorously validated, fine-grained lens for dissecting both expert and novice problem-solving strategies. After thorough investigation, we find that the thinking process of LRMs can be well-aligned with the episodes in the theory, as they also follow similar problem-solving processes. Thus, in this project, we aim to annotate the model (solver) responses with these episode categories. However, to better apply the theory to the analysis of model responses, we apply a hierarchical annotation structure, including paragraph-level annotation and sentence-level annotation. Paragraph-level annotation is used to capture the overall flow of the model response, including \textbf{three} categories: General, Explore, and Verify. Sentence-level annotation is used to capture the fine-grained behavior of each sentence, including \textbf{seven} categories: Read, Analyze, Plan, Implement, Explore, Verify, and Monitor.

This guidebook provides guidelines for \textbf{sentence-level annotation}. For the given model response sentences, each of them will be annotated as one of seven fine-grained episode types: Read, Analyze, Plan, Implement, Explore, Verify, or Monitor. The sentence-level annotation is used to capture the fine-grained behavior of each sentence, it depends on both the current sentence itself and its context.

\subsection*{\textbf{Label Definitions and Guidelines}}

\subsubsection*{\textbf{1. Read}}

\begin{itemize}[leftmargin=*]
    \item \textbf{Definition:} This is usually the initial phase, which focuses on extracting or restating the given information, conditions, and the goal of the problem as presented in the problem. It involves understanding the question without any inference of strategy or reasoning.
    \item \textbf{Guidelines:}
    \begin{itemize}[leftmargin=*]
        \item Sentences in this category should directly present the content of the original problem statement.
        \item Look for phrases that recall or repeat elements of the question.
        \item This label is mostly presented for the model's initial processing of the problem.
    \end{itemize}
\end{itemize}

\textbf{Keywords/Indicators:} ``The question asks...'', ``The problem requires...'', ``We are given...'', ``The goal is to...'', ``The choices are…'' direct quotes from the problem.

\textbf{Distinguishing Features:} This stage is purely about understanding the input, not about processing it or deciding how to solve it. Avoid labeling sentences as Read if they include any form of analysis or evaluation of the problem. The read stage usually appears at the beginning of the reasoning. However, it also appears in the middle of the reasoning, in order to ensure that the question was understood correctly.

\textbf{Example:} ``The question asks us to find the value of x in the equation \(2x + 5 = 10\).''

\subsubsection*{\textbf{2. Analyze}}

\begin{itemize}[leftmargin=*]
    \item \textbf{Definition:} This stage involves constructing or recalling relevant theories, introducing necessary symbols, and deducing relationships based on the problem statement and existing knowledge. The core activity is explanation or logical inference that sets the stage for the solution but does not involve concrete calculations yet.
    \item \textbf{Guidelines:}
    \begin{itemize}[leftmargin=*]
        \item Sentences should explain the underlying mathematical concepts or principles relevant to the problem.
        \item Look for the introduction of variables, formulas, or theorems.
        \item This label applies to logical deductions and inferences made with certainty.
    \end{itemize}
\end{itemize}

\textbf{Keywords/Indicators:} ``According to...'', ``We can define...'', ``This implies that...'', ``Therefore...'', ``Based on this...'', ``Let's denote...'', ``We can infer that...'', ``Let's note that ...'', ``Let me observe that ...'', ``Let's recall that ...''

\textbf{Distinguishing Features:} Analyze episode involves certain inferences and explanations, unlike Explore, which shows uncertainty. It usually precedes the actual execution of calculations in the Implement stage. Analyze does not involve any concrete calculation, which is unlike Implement.

\textbf{Important Note:} Be careful not to include sentences that involve substituting values or performing calculations, as those belong to the Implement stage.

\textbf{Example:} ``According to the Pythagorean theorem, in a right-angled triangle, the square of the hypotenuse is equal to the sum of the squares of the other two sides.'' or ``If I can get the equation in slope-intercept form \((y = mx + b)\), then I can plug in \(y = 4\) and solve for x, which should be d.''

\subsubsection*{\textbf{3. Plan}}

\begin{itemize}[leftmargin=*]
    \item \textbf{Definition:} This stage involves announcing the next step or outlining the entire solution strategy. It represents a commitment to a particular course of action before the actual execution begins.
    \item \textbf{Guidelines:}
    \begin{itemize}[leftmargin=*]
        \item Sentences should clearly state the intended next step or the overall plan.
        \item Look for explicit declarations of intent, often using the first person or imperative voice.
        \item This stage signifies that a decision has been made on how to proceed, and the next step should be related to math problem solving, rather than generally saying ``let's think about it''.
    \end{itemize}
\end{itemize}

\textbf{Keywords/Indicators:} ``Next, we will...'', ``The next step is to...'', ``We need to...'', ``Let's proceed by...'', ``I will now...'', ``The plan is to...'', ``We should first...'', ``To…, do…'', ``The xxx we need/want is…'', ``Let's…'', ``Then/Now calculate/consider ...''.

\textbf{Distinguishing Features:} The Plan phase clearly indicates the intended action, unlike Analyze, which explains concepts, or Explore, which suggests possibilities. It precedes the actual carrying out of the plan in the Implement stage. Note that sentences like ``Let's denote…'' is Analyze, because this is introducing a new variable, rather than making a plan. Sentences like ``let's verify…'' or ``let's double-check'' are Verify.

\textbf{Example:} ``Next, we will differentiate both sides of the equation with respect to x.''

\subsubsection*{\textbf{4. Implement}}

\begin{itemize}[leftmargin=*]
    \item \textbf{Definition:} This stage is the operational phase where the planned strategy is executed. It involves performing specific calculations, constructing diagrams, enumerating possibilities, or coding solutions using numerical values, symbols, or geometric objects.
    \item \textbf{Guidelines:}
    \begin{itemize}[leftmargin=*]
        \item Sentences should describe the actual steps taken to solve the problem.
        \item Look for mathematical operations, substitutions, and the generation of intermediate results.
        \item This stage is about ``doing'' the math.
    \end{itemize}
\end{itemize}

\textbf{Keywords/Indicators:} ``Substituting \(x = 2\), we get...'', ``Therefore, \(P(1) = -1\)'', ``Expanding the expression...'', ``The matrix becomes...'', actual mathematical equations and calculations.

\textbf{Distinguishing Features:} Implement involves concrete actions and calculations, unlike Analyze, which focuses on theoretical explanations, or Plan, which outlines future actions. If a conclusion follows the Implementation of math, that conclusion is tagged as Implement, such as ``therefore, the sum of all possible values is 5''.

\textbf{Example:} ``Substituting \(x = 3\) into the equation, we get \(2(3) + 5 = 6 + 5 = 11\).''

\subsubsection*{\textbf{5. Explore}}

\begin{itemize}[leftmargin=*]
    \item \textbf{Definition:} This stage is characterized by generating potential ideas, making guesses, drawing analogies, or attempting trial calculations that might be abandoned later. The model is exploring different avenues without committing to a specific solution path. This stage often involves uncertainty.
    \item \textbf{Guidelines:}
    \begin{itemize}[leftmargin=*]
        \item Sentences should suggest alternative approaches or possibilities.
        \item Look for tentative language and expressions of uncertainty.
        \item This stage involves brainstorming and initial investigations without a clear commitment to a particular method.
    \end{itemize}
\end{itemize}

\textbf{Keywords/Indicators:} ``Maybe we can try...'', ``Perhaps we could use...'', ``What if we consider...'', ``Another possibility is...'', ``Could this be related to...'', ``Maybe I should...'', ``Maybe there is another way...'', ``Maybe we can try ...'', ``Maybe there is a better way ...'', ``Maybe consider ...'', ``Perhaps ... is ...'', ``Let's try ...'', ``Alternatively, maybe use ...'', ``Wait, but maybe...'', ``But in soccer, it's possible to lose a game but still have more total goals?'' question marks indicating uncertainty about a step.

\textbf{Distinguishing Features:} Explore is marked by uncertainty and a lack of commitment, unlike Plan, which announces a definite course of action. It involves considering various options before settling on a specific plan. If a sentence contains analyzing the problem, implementing the calculation, or verifying the result or thought, even if it follows sentences like ``Maybe we can try …'', the sentences are not considered Explore at the sentence level, and therefore should not label them as Explore. Rather, these sentences are considered Analyze, Implement, or Verify within the Explore Episode at the paragraph level. Only the sentence like ``Maybe we can try …'' will be labeled as Explore at the sentence level.

\textbf{Example:} ``Maybe we can try substituting different values for x to see if we can find a pattern.''

\subsubsection*{\textbf{6. Verify}}

\begin{itemize}[leftmargin=*]
    \item \textbf{Definition:} This stage involves judging the correctness, effectiveness, or simplicity of the obtained result or the method used. It might include checking the answer, using an alternative method for calculation, or estimating bounds.
    \item \textbf{Guidelines:}
    \begin{itemize}[leftmargin=*]
        \item Sentences should express an evaluation or confirmation of the solution or the process.
        \item Look for keywords related to checking, confirming, or validating.
        \item This stage ensures the solution and result is accurate and makes sense.
    \end{itemize}
\end{itemize}

\textbf{Keywords/Indicators:} ``Let me double-check...'', ``This is consistent with...'', ``Plugging it back in...'', ``Therefore, the answer is correct.'', ``Let's confirm...'', ``Let me check again..'', ``We can confirm this by...'', ``This result seems reasonable because...'', ``The answer is ...?'', ``Is the answer ...?'', ``Is there any mistakes?'', ``Do I made mistakes?'', ``This is the same/correlated as previous ...'', ``But there seems contradict to ...'', ``... lead/arrive to the same answer'', ``Wait, we don't know ... yet'', ``Let's try another way to verify ...'', ``XXX is possible/impossible''. When the following sentences are to mean conclusions, ``... is indeed ...'', ``... should be...'', they are also at the Verify state.

\textbf{Distinguishing Features:} Verify focuses on evaluating the solution, unlike Implement, which focuses on generating it. It often involves comparing the result with initial conditions or using alternative methods.

\textbf{Example:} ``Let me double-check my calculations: \(2 \times 3 + 5 = 11\), which matches the previous result.''

\subsubsection*{\textbf{7. Monitor}}

\begin{itemize}[leftmargin=*]
    \item \textbf{Definition:} This additional category captures sentences that are typically short interjections or expressions indicating the model's self-monitoring, hesitation, or reflection at the juncture between different episodes. These often do not contain substantial problem-solving content and are brief pauses in the thought process.
    \item \textbf{Guidelines:}
    \begin{itemize}[leftmargin=*]
        \item Sentences should be short phrases indicating a shift in thought or a brief pause.
        \item Look for expressions of uncertainty, reflection, or transition.
        \item This label is for meta-comments that don't fit neatly into the other problem-solving stages.
    \end{itemize}
\end{itemize}

\textbf{Keywords/Indicators:} ``Hmm...'', ``Wait...'', ``Let me think.'', ``Okay...'', ``Let's see.'', ``Hold on.'', ``Let's see.'', ``But wait, hold on.'', ``Let me think.''

\textbf{Distinguishing Features:} Monitor sentences lack the substantive content of the other categories and primarily serve as indicators of the model's internal processing flow. They are often very short and act as bridges between more content-heavy stages.

\textbf{Example:} ``Wait.''

\subsection*{\textbf{Important Considerations for Annotators}}

\begin{itemize}[leftmargin=*]
    \item \textbf{Sentence-Level Focus:} Annotate each sentence individually based on its primary function within the problem-solving process.
    \item \textbf{Context is Key:} While keywords can be helpful, always consider the context of the sentence within the overall response. A sentence might contain a keyword but function differently based on the surrounding text.
    \item \textbf{Refer to Examples:} The examples provided in this guidebook and any additional examples you encounter should serve as valuable references.
\end{itemize}

\subsection*{\textbf{The Relation between Paragraph-level and Sentence-level Annotation}}

Although they focus on different granularities, paragraph-level and sentence-level annotations serve mutually complementary roles. There are three categories in paragraph-level annotation: General, Explore, and Verify. All three categories represent behaviors that may span large segments of responses and encompass several more fine-grained sentence-level categories. For example, during the initial attempt (General) at solving a problem, the solver can still exhibit other behaviors such as Plan, Explore, Verify, etc. Similarly, in the Explore and Verify phases, the solver can also demonstrate other behaviors like Plan, Explore, Verify, etc. This long-spanning characteristic motivates our use of a hierarchical structure. By applying this hierarchical structure, for sentences functioning like ``planning during the verification process,'' we do not need to struggle with which category to assign at the sentence level—Plan or Verify. We can directly use a paragraph-level annotation of Verify and sentence-level annotation of Plan to represent the behavior of this sentence.

\subsection*{\textbf{Some points that are easily misunderstood:}}

\begin{itemize}[leftmargin=*]
    \item ``Let me parse that information first.'' ``Let me do that.'' \\
    \textbf{should be Monitor}, rather than Plan. Because the solver is pausing to check understanding of the information already given, not outlining a future course of action.

    \item ``Wait, so if both lines pass through \((4, 1)\), that means that \((4, 1)\) is their point of intersection, right?'' \\
    \textbf{should be Explore}, rather than Analyze. Because the speaker is tentatively proposing an idea and seeking confirmation, not asserting a concluded analysis. The sentence shows uncertainty by the question mark.

    \item ``Hmm, but let me check again because that seems straightforward, but maybe there's a trick here.'' ``Wait, let me make sure.'' \\
    \textbf{should be Verify}, rather than Explore. Because ``let me check again'' explicitly marks a correctness check; the mention of a possible trick simply motivates the verification.

    \item ``Then it describes two lines.'' ``So, if both lines pass through \((4, 1)\), then that must be the intersection point.'' \\
    \textbf{should be Analyze}, rather than Explore. Because it summarizes structural information (categorizing the description) rather than hypothesizing or probing. No uncertainty is shown by maybe or perhaps.

    \item ``Wait, perhaps the description is a bit confusing.'' \\
    \textbf{should be Monitor}, rather than Explore. Because it reflects on the clarity of understanding (metacognitive monitoring).

    \item ``The slope \(m_1\) is \((0 - 1)/(32/7 - 4)\).'' ``So equation of first line is \(y = -7/4 x + 8\).'' \\
    \textbf{should be Analyze}, rather than Implement. Because it is performing the analytic calculation (finding slope), not carrying out a pre-defined multi-step procedure.

    \item ``Wait, \(7/4\) times 4 is 7.'' \\
    \textbf{should be Analyze}, rather than Verify. Because it is a micro-check of arithmetic inside the working process (self-monitoring), not a separate verification stage.

    \item ``Now, the second line passes through \((32/9, 0)\) and \((4, 1)\).'' \\
    \textbf{should be Analyze}, rather than Read. Because it sets up data for further calculations on the second line (transition into analysis).
\end{itemize}

\end{document}